\documentclass[letterpaper, 10 pt, conference]{ieeeconf}  

\IEEEoverridecommandlockouts             

\usepackage{graphicx}
\usepackage{subfigure}
\usepackage[ruled,linesnumbered]{algorithm2e}
\usepackage{algpseudocode}
\usepackage{multirow}
\usepackage{multicol}
\usepackage{makecell}
\usepackage{textcomp}
\usepackage{hyperref}
\usepackage{stfloats}
\usepackage{cite}
\usepackage{amsfonts,amssymb}
\usepackage{amsmath}

\title{\LARGE \bf
HPHS: Hierarchical Planning based on Hybrid Frontier Sampling for Unknown Environments Exploration
}

\author{Shijun Long\textsuperscript{1}, Ying Li\textsuperscript{1}, Chenming Wu\textsuperscript{2}, Bin Xu\textsuperscript{1}, and Wei Fan\textsuperscript{1} 
\thanks{This work was supported in part by the National Natural Science Foundation of China under Grant 52102449, in part by the China Postdoctoral Science Foundation under Grant 2021M690394, and in part by the Beijing Institute of Technology Research Fund Program for Young Scholars and S\&T Program of Hebei under Grant 21567606H.}
\thanks{$^{1}$S. Long, Y. Li, B. Xu, and F. Wei are with the School of Mechanical Engineering, Beijing Institute of Technology, Beijing, China. {\tt\small \{sj\_long, ying.li, bitxubin, fanweixx\}@bit.edu.cn}}%
\thanks{$^{2}$C. Wu is with RAL, Baidu Research. {\tt\small wuchenming@baidu.com}}
\thanks{$^{1}$Available at {\tt\small https://github.com/bit-lsj/HPHS.git}}%
}

\begin{document}

\maketitle
\thispagestyle{empty}
\pagestyle{empty}

\begin{abstract}
Rapid sampling from the environment to acquire available frontier points and timely incorporating them into subsequent planning to reduce fragmented regions are critical to improve the efficiency of autonomous exploration. We propose HPHS, a fast and effective method for the autonomous exploration of unknown environments. In this work, we efficiently sample frontier points directly from the LiDAR data and the local map around the robot, while exploiting a hierarchical planning strategy to provide the robot with a global perspective. The hierarchical planning framework divides the updated environment into multiple subregions and arranges the order of access to them by considering the overall revenue of the global path. The combination of the hybrid frontier sampling method and hierarchical planning strategy reduces the complexity of the planning problem and mitigates the issue of region remnants during the exploration process. Detailed simulation and real-world experiments demonstrate the effectiveness and efficiency of our approach in various aspects. The source code will be released to benefit the further research$^{1}$.
\end{abstract}

\section{Introduction}

As the autonomous ability of robots has been constantly improved, increasing robots are being used for rescue, mapping, exploration, 3-D reconstruction, and other tasks. However, for various reasons, human operators are not always able to control the robot's movement and complete tasks in real time. Such a situation requires the robot to have higher autonomy and be able to independently decide where to go and complete the mapping of the environment. Therefore, more researches  \cite{yamauchi1997frontier,perkasa2020improved,image,sensor_reading,huang2023fael} aim at developing autonomous exploration techniques to improve the exploration efficiency of robots in unknown environments. 

The mainstream framework of exploration technology mainly includes frontier-based and sampling-based methods. Both methods essentially search the potentially unknown areas of the environment and select the high-revenue one to explore. The boundary of the unknown area is expanded continuously until the whole environment is completely modeled. However, these methods still have some limitations:
\begin{itemize}
\item Expensive computation. Most methods need to search the boundary of the unknown environment in a large area or perform a large number of frontier sampling, which is a huge burden for edge computing devices and causes the failure of the robot to respond quickly \cite{huang2023fael}. When the exploration range increases, the efficiency of sampling becomes lower, resulting in the inability to generate new frontiers in real time.
\item Greedy strategy. Common revenue evaluation methods either minimize the distance from the robot to the frontier \cite{yamauchi1997frontier}, or maximize the information gain \cite{gao2018improved}. However, they all ignore efficient global path coverage, resulting in myopic behaviors of the robot. This leads to backtracking exploration and low efficiency.
\end{itemize}

\begin{figure}[!t]\centering
	\includegraphics[width=7.2cm]{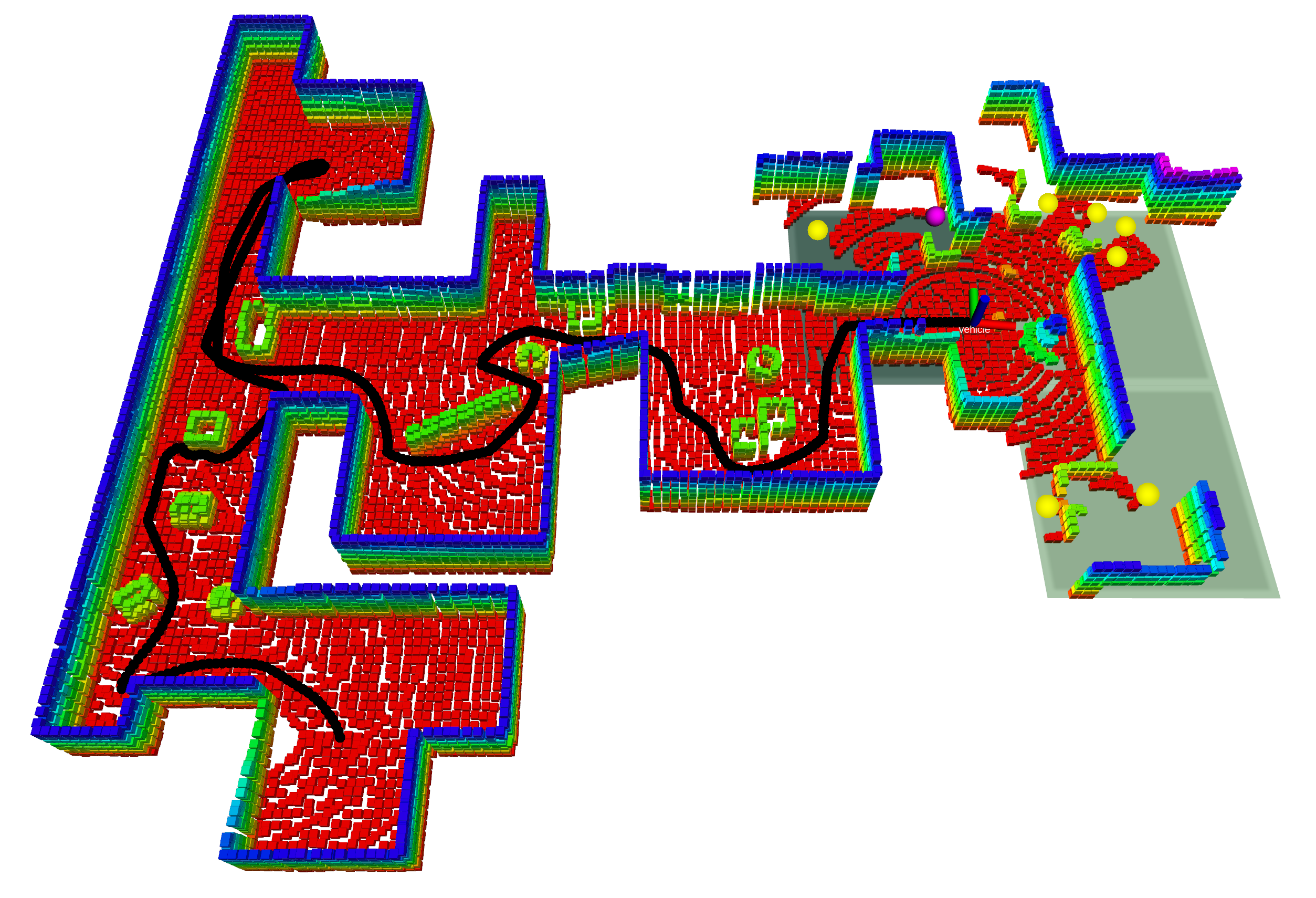}
	\caption{A robot performs autonomous exploration in an unknown environment using the proposed method. The yellow points represent the frontier points, the purple point is the next target point, and the black curve is the trajectory of the robot. The green grids are divided subregions.}
    \label{effect}
\end{figure}

\begin{figure}[!t]\centering
    \resizebox{\linewidth}{!}{
	\includegraphics[width=7.5cm]{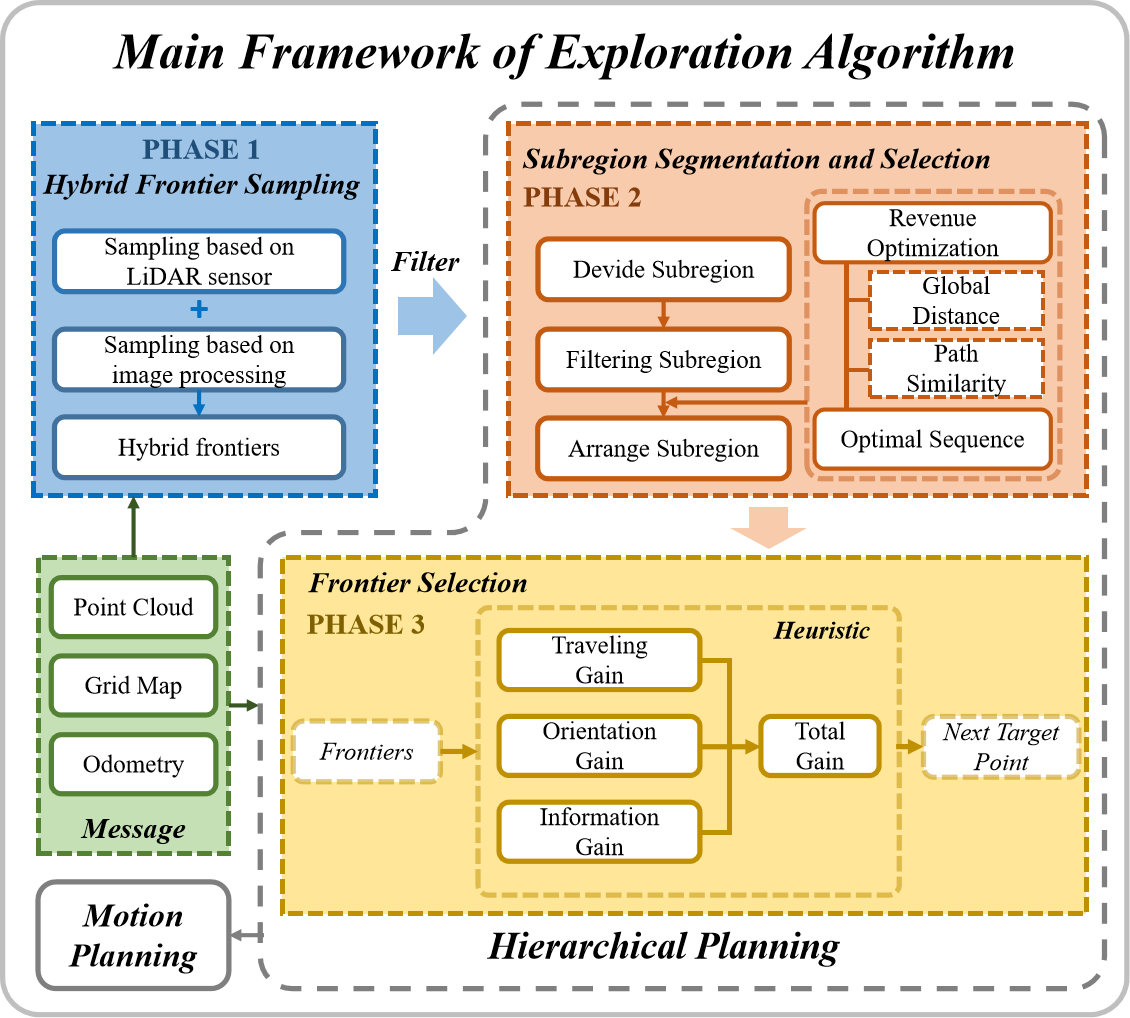}
 }
	\caption{The overall framework diagram of the exploration system.}
    \label{framework}
\end{figure}

In this paper, we present a hierarchical planning exploration system based on hybrid frontier sampling. Firstly, the hybrid frontier sampling method is proposed to quickly extract the frontier points by directly using LiDAR data and local map information. Then, the hierarchical planning strategy divides the updated environment into multiple subregions, reorders the exploring sequence of subregions, and evaluates the benefits of frontier points within the currently accessed subregion. This two-level planning strategy enables the robot to explore regions in a sequence determined by both global and local information. Finally, the goal frontier point selected from the subregion is taken as the next target to guide the robot for further exploration. The whole process is repeated until the environment is fully modeled.

We evaluate the proposed method in both simulation and real environments. The results show that the proposed method has an excellent performance in terms of exploration efficiency and completion. The main contributions of this paper are as follows:

1) A fast and efficient hybrid frontier sampling method is proposed to extract frontier points, ensuring that the sampling efficiency is not affected by the scale of the whole environment map.

2) The hierarchical planning strategy is adopted to provide global information for the robot and reduce the complexity of the optimization problem.

3) Extensive experiments are conducted to verify the theoretical and practical feasibility of the proposed method, and our method shows more advantages compared with other methods in terms of traveling length and exploration time.

\section{Related Works}
Environmental exploration methods are mainly divided into frontier-based and sampling-based methods. With the development of deep learning, there are also methods to use these learning-based technologies to enable autonomous exploration \cite{deep1,deep2,deep3,deep4,deep5}.

The frontier-based method is first proposed by Yamauchi et al. \cite{yamauchi1997frontier}, by extracting the boundary between known and unknown regions as the frontier, and selecting the frontier closest to the robot as the next exploration target area. This approach can easily promote the exploration task into a local optimal situation, causing the robot to move to the next area before it has explored one, or repeatedly explore the environment. Although there are subsequent works to introduce information gain into the evaluation function \cite{perkasa2020improved}, the above problem cannot be well solved by simply adding an evaluation indicator. Different from conventional methods to obtain frontiers, \cite{image} uses image processing technology to extract frontiers in the environment map, but with the expansion of exploration scope, processing this map will consume more and more computing resources. To reduce the resource consumption of generating frontiers, \cite{sensor_reading} uses laser reading to detect the frontiers, which speeds up the generation of frontiers. The approach in \cite{holz2010evaluating} uses the map segmentation method to divide each indoor room and corridor into subregions to reduce the problem of multiple visits to the environment. Meanwhile, repeated inspection technology is used to improve the efficiency of the robot in the exploration process.

The Next Best View Planner \cite{nbvp} is a representative approach of the sampling-based method, which uses RRT to expand the entire exploration space, selecting the branches of the tree with the most revenue for the robot to advance. Since the exploration work itself lacks information on the global environment, the robot cannot predict the future environmental space, so the planned path is difficult to be globally optimal. To provide the robot with a global perspective, \cite{tdle} divides the environmental map into different areas and makes global path planning for these areas to provide global guidance for the exploration system. \cite{tare} employs a hierarchical framework to simultaneously process online update environment representations and search for continuous traversable paths, significantly enhancing exploration efficiency in large-scale environments. To speed up the sampling process, \cite{dsvp} utilizes the dynamic expansion of a two-stage planning method, rather than rebuilding the tree, only part of the tree nodes are generated in each iteration process, which greatly improves the computational efficiency. 

Some of the above methods use RRT to sample frontier points or viewpoints. However, due to the randomness of the RRT algorithm, the generated nodes may not be uniform. As the tree grows, the generation of new nodes becomes slower, and some small areas and passages are difficult to sample, resulting in incomplete exploration. Several methods \cite{fuel, huang2023fael} adopt the incremental detection to extract frontiers, but require maintaining a state voxel map at the underlying level to search frontiers. This consumes computation memory when the scale of the map becomes larger.

In our work, we quickly obtain the frontier points of the surrounding environment, the frontier points are extracted directly from the LiDAR point cloud and the local map, which avoids traversing and maintaining a large-scale voxel map. We also adopt the hierarchical planning strategy to provide comprehensive information for the exploration system, this can make efficient decisions while further reducing the computing consumption.

\section{Proposed Method}

The exploration system consists of four modules: the hybrid frontier point sampling module, the filter module, the subregion segmentation and selection module, and the frontier selection module. The hybrid frontier point sampling module is used to quickly sample potential frontier points from the environment. The filter module removes some invalid frontier points. The subregion segmentation and selection module is proposed to divide the region of the map into several subregions and arrange their access order, then the appropriate subregion is assigned to the next module. According to the assigned subregion, the frontier selection module sorts the candidate frontier points in the subregion using the heuristic evaluation criteria. The point with the highest gain is chosen as the next marching target. The overall framework diagram of the exploration system is shown in Fig. \ref{framework}.

\begin{figure}[!t]\centering
	\includegraphics[width=6cm]{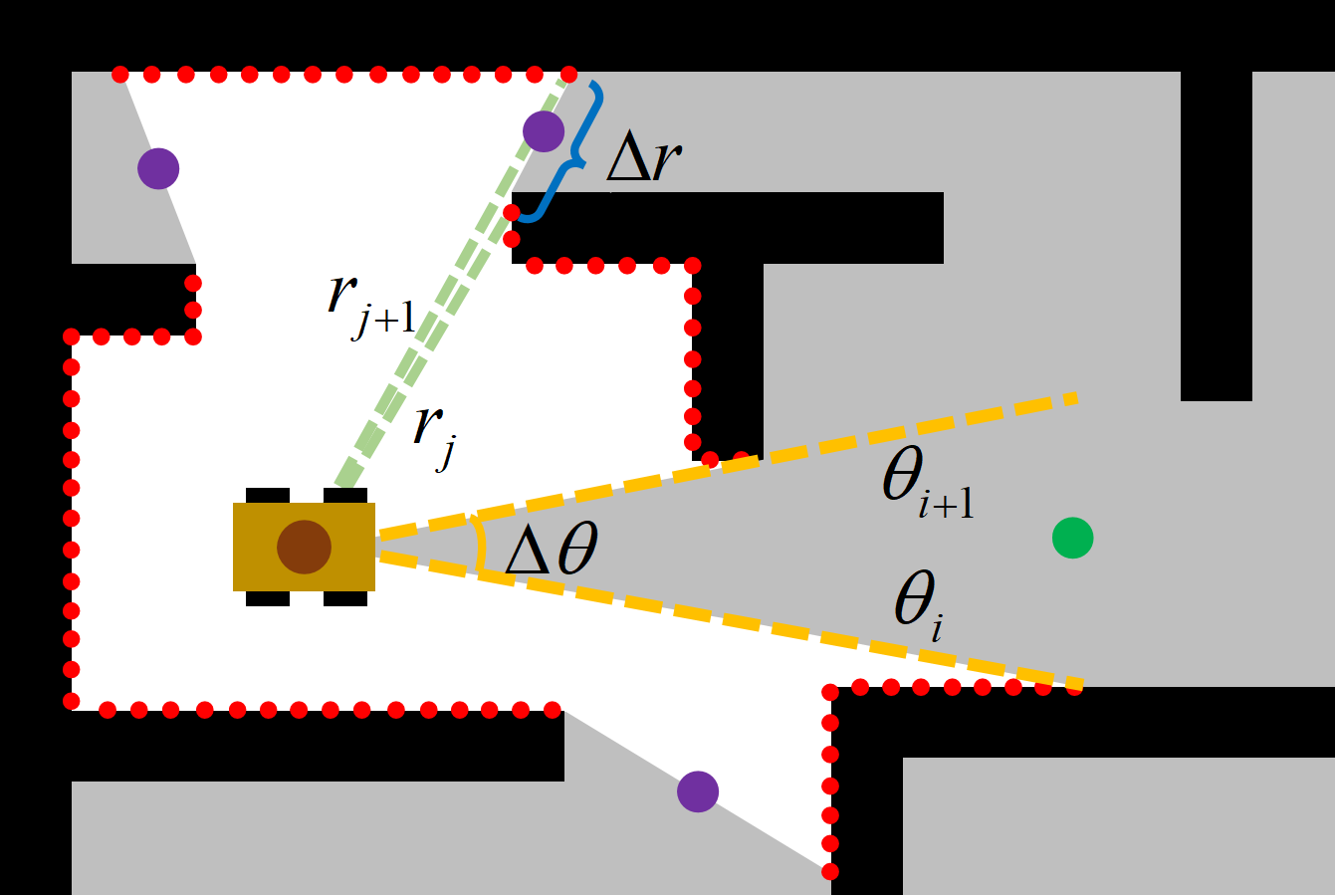}
	\caption{Frontier points sampling directly from the LiDAR data. The red dots represent the point cloud. The purple and green dots represent the inserted frontier points when condition 1 or 2 is satisfied respectively. The radius and polar angle of the point in the polar coordinate system are expressed by $r$ and $\theta$.}
    \label{frontier_sensor}
\end{figure}

\subsection{Hybrid Frontier Point Sampling Module} Fast extraction of frontier points from the environment is essential to improve exploration efficiency. We adopt a hybrid frontier sampling method. Specifically, we first determine the potential frontier points directly from the LiDAR point cloud, and further screen them to obtain the final available frontier points $\mathcal{F}_{sensor}$. With the movement of the robot in the environment and updating of the LiDAR data, new frontier points will be generated to help expand the boundary. Due to the presence of environmental noise, some frontier points may not be detected. To obtain frontier points stably, we also introduce the image processing-based frontier detector to acquire frontier points $\mathcal{F}_{image}$ within a radius of $d_s$ around the robot. Finally, the frontier points $\mathcal{F}_{all}$ obtained by two frontier detection methods are gathered together.
  
\textit{1) Sampling frontier points based on LiDAR sensor}

We are inspired by the method in \cite{gdae} for its several advantages such as fast detection speed and long detection distance. In the frontier point sampling stage, the point cloud of the current frame is converted to the cylindrical coordinate system. Then, the point cloud is selected based on a specific height $z$, and sorted according to the polar angle from $0^{\circ}$-$360^{\circ}$ to construct a collection of point cloud $\mathcal{L}^z=\left\{l_{\theta_0}^z,l_{\theta_1}^z,...,l_{\theta_n}^z\right\}$. For any two consecutive points $l_{\theta_i}^z$, $l_{\theta_{i+1}}^z$ in the collection, a frontier point is inserted between them if satisfy one of the following conditions in the Eq. (\ref{condition1})-(\ref{condition2}).

\begin{equation}
\vert r_{i+1}-r_i \vert \geq r_{gap} \;\;or \label{condition1}
\end{equation}

\begin{equation}
\theta_{i+1}-\theta_i \geq \theta_{inf} \label{condition2}
\end{equation}
where $r$ and $\theta$ represent the radius and polar angle of the point cloud in the polar coordinate system respectively. The condition (\ref{condition1}) is satisfied when the radius of two consecutive points differs by a certain distance, which means that there may be a corridor passage or slightly larger occlusions between them. Due to these occlusions, the robot is not able to complete the mapping in this area. The condition (\ref{condition2}) states that there is no scanning point in the region between $\theta_{i}$ and $\theta_{i+1}$, which reveals that the region has not been mapped either. Furthermore, setting a frontier point in this region can help the robot detect and expand the boundary, especially in exploring a large environment.

The presence of frontiers in the map is usually a result of obstructing or exceeding the maximum detection distance of the sensor. Therefore, points that meet these conditions are considered to exist within unexplored areas. Compared with searching the whole map or random sampling to obtain the frontier point, the direct use of LiDAR data can insert the frontier point more accurately and quickly without traversing the whole map, so the calculation speed is not affected by the scale of the map. Fig. \ref{frontier_sensor} intuitively represents the above two approaches to sampling frontier points.

\textit{2) Sampling frontier points based on image processing}

Not all frontier points can be found using sensor information only, thus we complement the frontiers with an image processing-based frontier detector \cite{image}. This detector leverages edge and contour detection methods to process the map data in order to capture the intersection of known and unknown areas. Image-based detection algorithms consume more computing resources with the increase in the image size. Our method aims to reduce the computation burden and improve the time efficiency when extracting frontier points. Thus, we restrict the scale of the local map for detecting frontier points within a radius of $d_s$.

\subsection{Filter Module}

The filter module accepts all sampled frontier points $\mathcal{F}_{all}$ for filtering to obtain available points. A sampled frontier point will be deleted if it is located near obstacles or other frontier points already exist within its field of view. Besides, this filter also rejects points that belong to the known region. The final filtered frontier points ${\mathcal{F}_{filt}}$  are used for the further exploration.

Additionally, the filter module also deletes frontier points that the robot has visited or unreachable frontier points in real time.

\begin{figure*}[htbp]
\centering
\includegraphics[width=16cm]{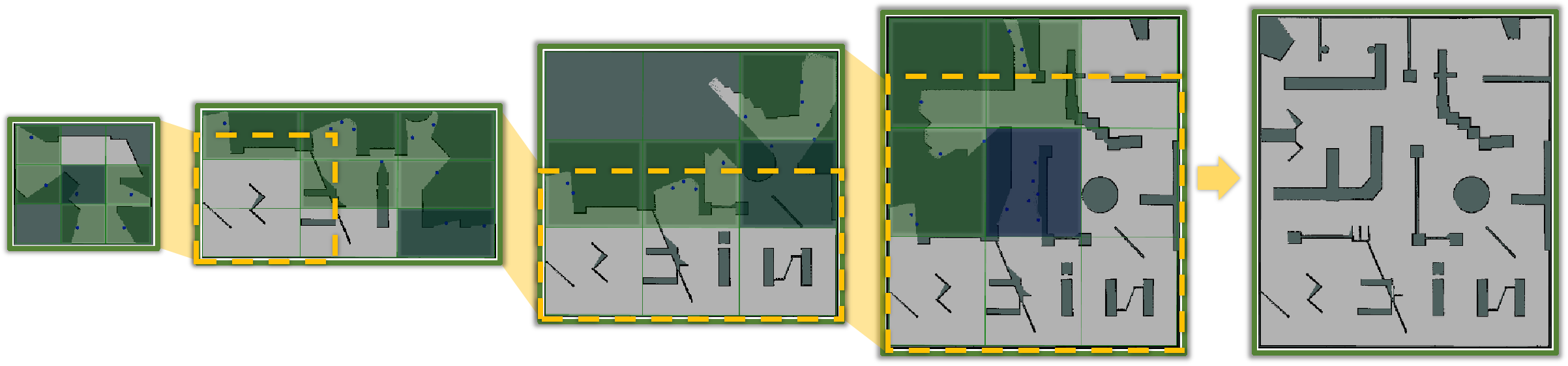}
\caption{Subregion segmentation and selection during the exploration. The green grids represent the remaining subregions after being filtered, and the dark blue grid in each map refers to the subregion that should be visited at the current moment. The other unfilled grids represent subregions that are filtered out, as there are no frontiers in the interior. As the size of the map changes, subregions should grow dynamically.}
\label{subregion}
\end{figure*}

\subsection{Subregion Segmentation and Selection Module}

This module processes the whole map region $\mathcal{M}$ according to the pipeline of dividing subregions, filtering, and arranging the access order of subregions. The boundary $\mathcal{B}$ of the current map is first detected to determine the maximal rectangular bounding box $\mathcal{R}$. Then, the entire rectangle $\mathcal{R}$ with the size of $\mathcal{R}_w\times \mathcal{R}_h$ is split into several subregions uniformly in both height and width. Thus, the map is divided into $n$ subregions $\mathcal{SR}=\left\{sr_0,sr_1,...,sr_n\right\},n\leq {n_wn_h}$. The filtered frontier points $\mathcal{F}_{filt}$ are divided into their respective subregion $sr\in\left\{sr_0,sr_1,...,sr_n\right\}$, according to their positions. Those subregions without any frontier point are filtered out. Noting that the map boundary $\mathcal{B}$ is constantly updated as the map area $\mathcal{M}$ expands during the exploration process, the size and center position of each subregion will also dynamically change as the map is extending. Fig. \ref{subregion} explains the dynamic division of subregions throughout the exploration.

In each of the remaining subregions $\mathcal{SR}^{filt}=\left\{sr_0^{filt},sr_1^{filt},...,sr_m^{filt}\right\},m\leq n$, there exists at least one frontier point. When entering any of these subregions, the new environmental information can be obtained. In order to arrange the sequence of the exploration for the remaining subregions, that is, to find a global path covering the exploration space, we present a revenue function for the global path that simultaneously considers the global coverage and the traveling distance. The optimal subregion arrangement is found by optimizing the sequence combination of subregions by maximizing the revenue function:
\begin{equation}
    \begin{aligned}
        \mathbf{G}^* &= \max Rev\left(\mathbf{G}\right) \\
                     &=\max e^{-\lambda_2\cdot DTW(\mathbf{G})} \cdot\sum_{i=0}^m e^{-\lambda_1\cdot
\mathbf{D}\left(\mathcal{P}_{rob},\mathcal{P}_i\right)}
    \end{aligned}
    \label{revenue}
\end{equation}

\begin{equation}
    \begin{aligned}
    & \mathbf{D}\left(\mathcal{P}_{rob},\mathcal{P}_i\right) = \lambda_3 \cdot \mathbf{E}\left(\mathcal{P}_{rob},\mathcal{P}_0\right) + \mathbf{E}\left(\mathcal{P}_{0},\mathcal{P}_1\right) +\cdots \\
    & +\mathbf{E}\left(\mathcal{P}_{i-1},\mathcal{P}_i\right)
    \end{aligned}
    \label{dist}
\end{equation}
where $\mathbf{G}=[sr_0^{filt},sr_1^{filt},...,sr_m^{filt}]$ is the access sequence of the remaining subregions, $\mathbf{D}\left(\mathcal{P}_{rob},\mathcal{P}_i\right)$ is the cumulative distance from the robot position $\mathcal{P}_{rob}$ to the subregion center position $\mathcal{P}_i$, and $\lambda$ is the turning factor. The revenue function $Rev(\mathbf{G})$ in Eq. \ref{revenue} considers the path coverage length when passing through centers of these subregions, and introduces the Dynamic Time Warping (DTW) method \cite{dtw} to calculate the similarity between the global path $\mathbf{G}$ and the path sequence selected in the last iteration. In Eq. \ref{dist}, $\mathbf{E}\left(\mathcal{P}_{i-1},\mathcal{P}_i\right)$ represents the Euclidean distance between centers of two subregions, and the factor $\lambda_3$ is introduced to allow the exploration to start in the subregion that close to the robot. This prevents the robot from starting exploration from a distant region, reducing unnecessary backtracking.

Eq. \ref{revenue} is proposed to solve a combinatorial optimization problem, which essentially finds an optimal sequence from all possible permutation combinations. Additionally, Eq. \ref{revenue} ignores the amount of information gain, since all the remaining subregions are worth exploring and only the coverage distance of the global path needs to be considered. After arranging the sequence, the subregion that should be visited currently is selected according to the access sequence.

\subsection{Frontier Selection Module}
\begin{figure}[!t]\centering
	\includegraphics[width=6cm]{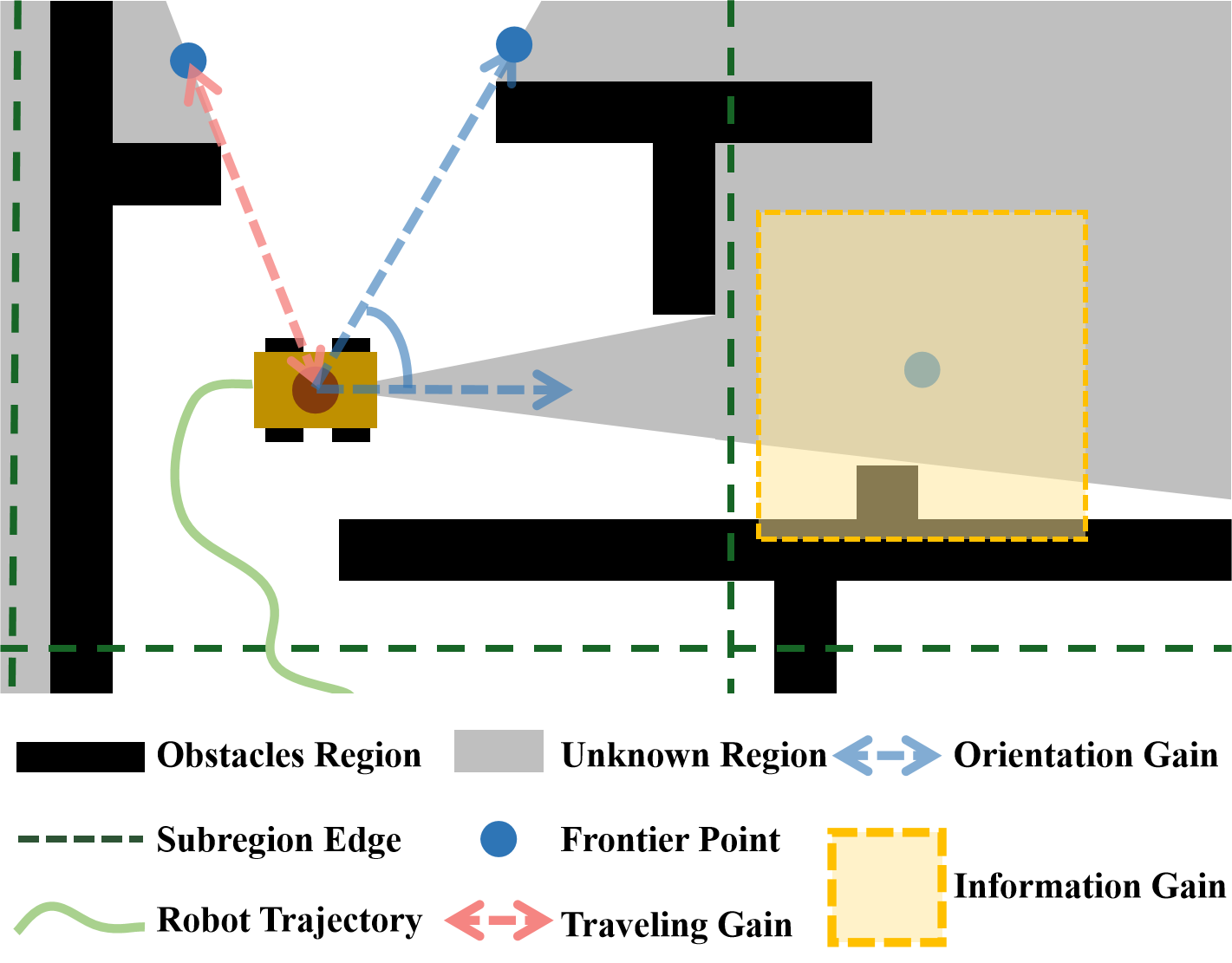}
	\caption{The illustration of calculating the heuristic gain. The total gain of a frontier point is composed of the traveling gain, the orientation gain, and the information gain. The information gain uses a $k\times k$ information kernel for the comprehensive evaluation.}
    \label{gain}
\end{figure}

This module receives the assigned subregion and selects the target point $target\in \left\{ \mathcal{F}^{sub}\right\}$ in the assigned subregion using the heuristic function. Then we assign the target point to the planning module to drive the robot to explore. Fig. \ref{gain} illustrates the calculation of the heuristic gain. The frontier selection module selects the best point by considering:

\textbf{Traveling gain:} The distance between the robot position $\mathcal{P}_{rob}$ and the frontier point $\mathcal{F}_j^{sub}$ in the assigned subregion reflects the travel cost that the robot needs to pay to reach the frontier point. To reduce the computational cost, we calculate this gain $G^{D}$ using the Euclidean distance:
\begin{equation}
G^{D}_j=\mathbf{E}\left(\mathcal{P}_{rob},\mathcal{F}_j^{sub}\right)
\end{equation}

\textbf{Orientation gain:} It is defined as the angle $\theta_{ori}$ between the current motion orientation of the robot and the vector between the robot and the frontier point $\mathcal{F}_j^{sub}$. The point with a small angle is beneficial to maintain the consistency of the robot's motion, avoiding taking a zigzag path or turning back to explore the point behind it. The orientation gain $G^{O}$ is defined as follows:
\begin{equation}
G^{O}_j=e^{2 \cdot \left(\frac{2 \theta_{o r i}}{\pi}-1\right)}
\end{equation}
 where larger  $\theta_{ori}$ values are more penalized, this can avoid unnecessary routes. It also prevents small values of  $\theta_{ori}$ from becoming the decisive factor, since it is reasonable to explore within a range of angles.

\textbf{Information gain:} The information gain refers to the revenue calculated from the search space $\mathbb{R}$ around the frontier point. Each grid in the map has three possible states: free $\mathbb{R}_f$, unknown $\mathbb{R}_u$, and occupied $\mathbb{R}_o$, i.e, $\mathbb{R}=\mathbb{R}_f\cup\mathbb{R}_u\cup\mathbb{R}_o$. The unknown state indicates that the area has not been explored, the free state means that the region has been explored and not occupied, and the occupied state represents that there exist obstacles that affect the safety of the robot. The integrated information gain $G^{I}$ for this region is calculated using a $k\times k$ information kernel:
\begin{equation}
G^{I}_j=e^{\frac{\sum_{w=-\frac{k}{2}}^{\frac{k}{2}} \sum_{h=-\frac{k}{2}}^{\frac{k}{2}} S_{w,h}}{k^2}}
\end{equation}
where $s$ is assigned according to the state of the grid in the map and should satisfy:
\begin{equation}
0\leq s_o\leq s_f\leq s_u
\end{equation}

Unlike the conventional information calculation method \cite{perkasa2020improved}, Eq. (5) also imposes penalties on other states, especially areas with obstacles. It is similar to the heuristic used in \cite{informed}. Our goal is to enable the robot to explore the entire environment as safely as possible.

\textbf{Total gain:} For any frontier point $\mathcal{F}_j^{sub}$ in the assigned subregion, the total heuristic gain $Gain$ can be calculated as:
\begin{equation}
Gain=\tau_3\cdot\Vert G^{I}_j \Vert - \tau_1\cdot\Vert G^{D}_j \Vert - \tau_2\cdot \Vert G^{O}_j \Vert
\end{equation}
where $\tau$ denotes the weight factor, and the min-max normalization is used to normalize indicators. The point with the largest score is selected as the next target to explore:
\begin{equation}
target=\arg\max _{\mathcal{F}^{sub}_j} Gain
\end{equation}

\subsection{Advantages of Hierarchical Strategy}

\textit{Improving decision speed.} A significant amount of computing resources will be consumed if evaluating all these frontier points in the map, resulting in a slow decision speed. By dividing the environment into several subregions, the large-scale exploration challenge can be decomposed into multiple smaller-scale problems, which reduces the complexity of the task.

\textit{Avoiding local minimum trap.} It is difficult to get a globally optimal solution by evaluating the frontier points distributed in different regions only by several evaluation metrics. This leads to unnecessary reentry and repeated exploration. The hierarchical planning strategy enables the robot to make decisions on a larger scale initially than just based on the current local information. Such operation helps prevent the robot from falling into a local minimum while striking a balance between the information acquisition and the exploration efficiency.

\textit{Quickly responding to environmental changes.} Hierarchical planning allows for a more convenient response to environmental changes and enables real-time updates to the exploration strategy. By managing and optimizing the exploration process through the division of regions, the exploration system becomes adaptable to various types and scales of environments.

\begin{algorithm}[!t]
    \caption{Exploration}
    \label{alg:Exploration}
    \KwIn{$\mathcal{P}_{rob}$: \textit{Robot position}, $\mathcal{M}$: \textit{2-D grid map}}
    \KwOut{$TargetPoint$}
        $Flag \gets True$\\
        $BestGain \gets 0$\\
        \While{$Flag=True$}{
            $\mathcal{F}_{all} \gets$ \texttt{SamplingFrontiers()}\\
            $\mathcal{F}_{filt} \gets$ \texttt{Filter($\mathcal{F}_{all}$)}\\
            $\mathcal{SR} \gets$ \texttt{SubregionSegmentation()}\\
            $\mathbf{G}^* \gets$ \texttt{ArrangeSubregions($\mathcal{SR}$)}\\
            Select the subregion visited at the current instant\\
            \For{\textit{j from} 0 \textit{to N}}{
                \If {\texttt{Gain($\mathcal{F}^{sub}_j)$}$>BestGain$}{
                    $BestGain \gets $\texttt{Gain($\mathcal{F}^{sub}_j$)}\\
                    $TargetPoint \gets \mathcal{F}^{sub}_j$\\
                }
            }
            \If{$\mathcal{F}_{all}=\emptyset$}{
                $Flag \gets False$\\
            }
        }
\end{algorithm}

\section{Simulation And Experiment}

\begin{figure}[!b]\centering
	\includegraphics[width=5cm]{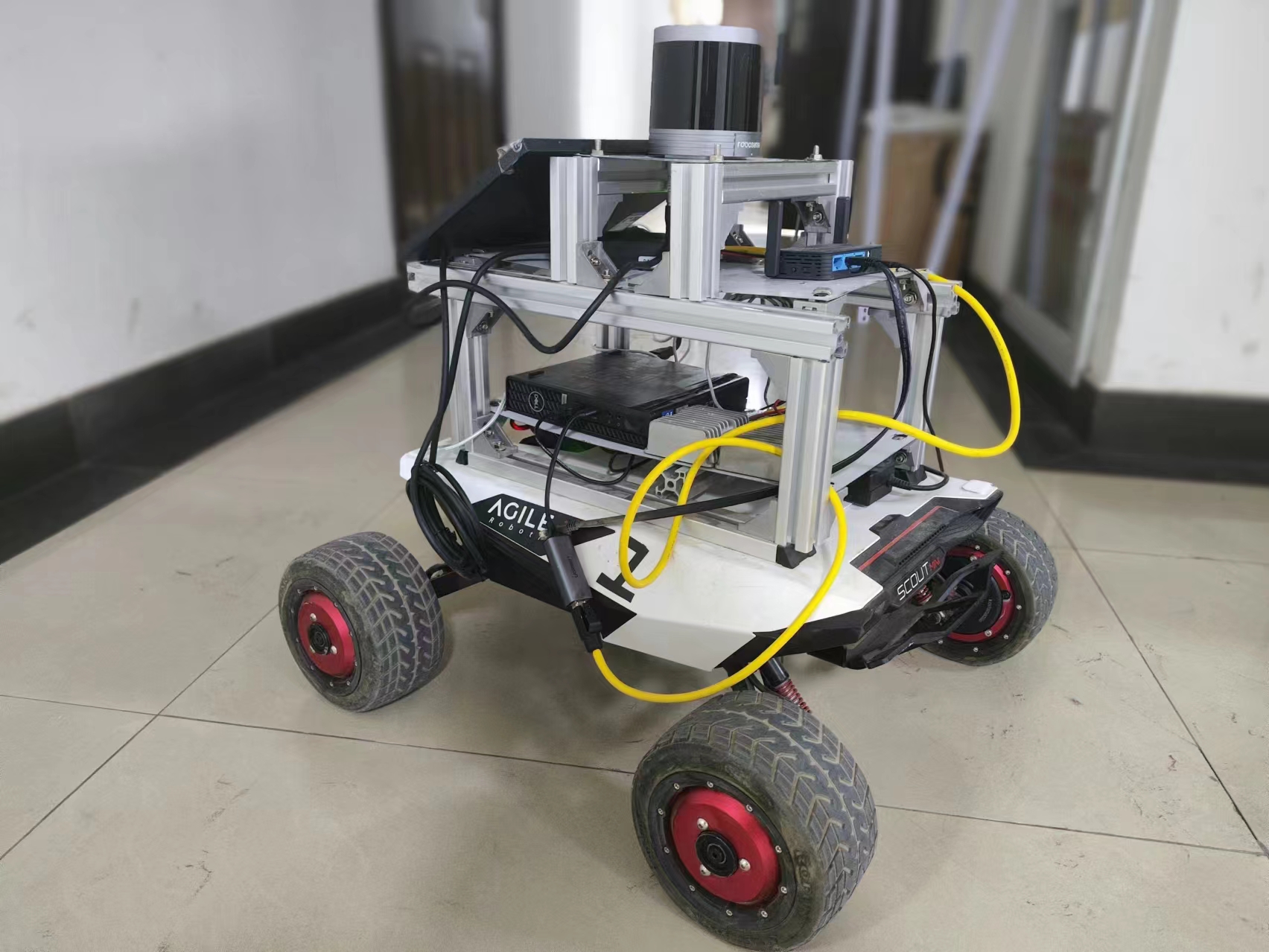}
	\caption{The real-world ground platform.}
    \label{platform}
\end{figure}

The proposed method is validated in both simulation and real-world experiments. Fig. \ref{platform} shows our real-world ground platform with an Intel Core i5-11500T@1.5 GHz CPU and 16 GB RAM using Ubuntu 18.04 LTS. The platform is topped with the RoboSense RS-Hellos-16P LiDAR sensor.

\subsection{Simulation Experiment}
We build four environments for simulation experiments, namely, maze, office, indoor 1, and indoor 2. The four scenarios have different structural arrangements to evaluate the exploration performance of the algorithm in each of them. In particular, two indoor environments are composed of a series of corridors and rooms, and there exist many dead corners that are not easy to explore. The robot is easy to miss some places during the exploration process, resulting in the formation of a backtracking path. All the scenes are built in the Gazebo simulator.
 
We evaluate the performance of our method by comparing it with three previous methods in these metrics: exploration time, exploration distance, exploration rate (the ratio of explored area to traveling distance), and exploration completion. 
\begin{itemize}
\item \textit{Efficient Dense Frontier Detection} \cite{frontier}: A frontier-based method that exploits the submap structure of the SLAM to quickly perform frontier updates and achieve responsive exploration goal planning.

\item \textit{TDLE} \cite{tdle}: An improved method based on RRT-exploration \cite{rrt_exploration}. It employs the regional division and arrangement to efficiently obtain a global view for exploration.

\item \textit{TARE} \cite{tare}: One of the state-of-the-art exploration methods that uses a hierarchical framework to represent the environment space, which can efficiently deal with large-scale complex environments and achieve faster exploration.

\end{itemize}
All the tests use the exploration algorithm as the top-level decision-making module, while the planning and obstacle avoidance module as the middle layer uses the open source framework \cite{cao2022autonomous} of the Robotics Institute from Carnegie Mellon University and obtains the metric data during the exploration process. In each scenario, simulation experiments are conducted 10 times with each method.

\begin{figure}[!t]
\centering
\subfigure[Maze]{
    \includegraphics[width=4cm]{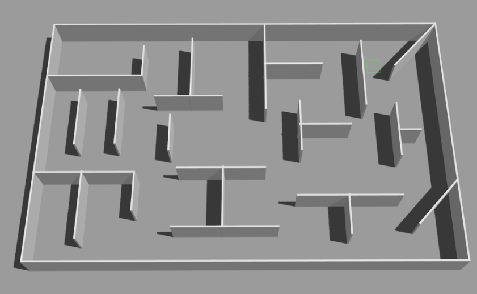}}
\subfigure[Office]{
    \includegraphics[width=4cm]{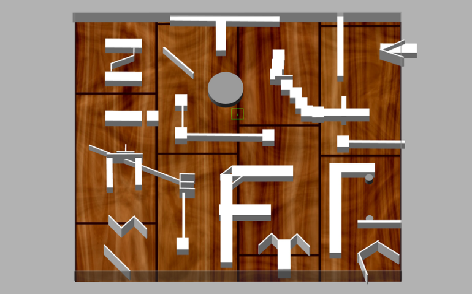}}
\\
\subfigure[Indoor 1]{
    \includegraphics[width=4cm]{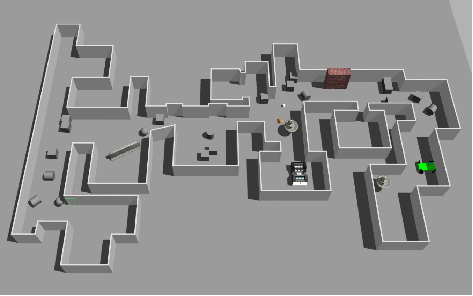}}
\subfigure[Indoor 2]{
    \includegraphics[width=4cm]{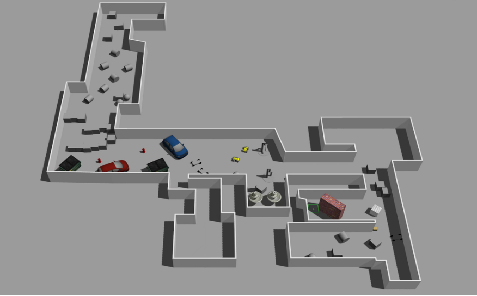}}
\caption{The four simulation environments.}
\label{scenes}
\end{figure}

\begin{figure}[!t]
\centering
\subfigure[Maze]{
    \includegraphics[width=6.5cm]{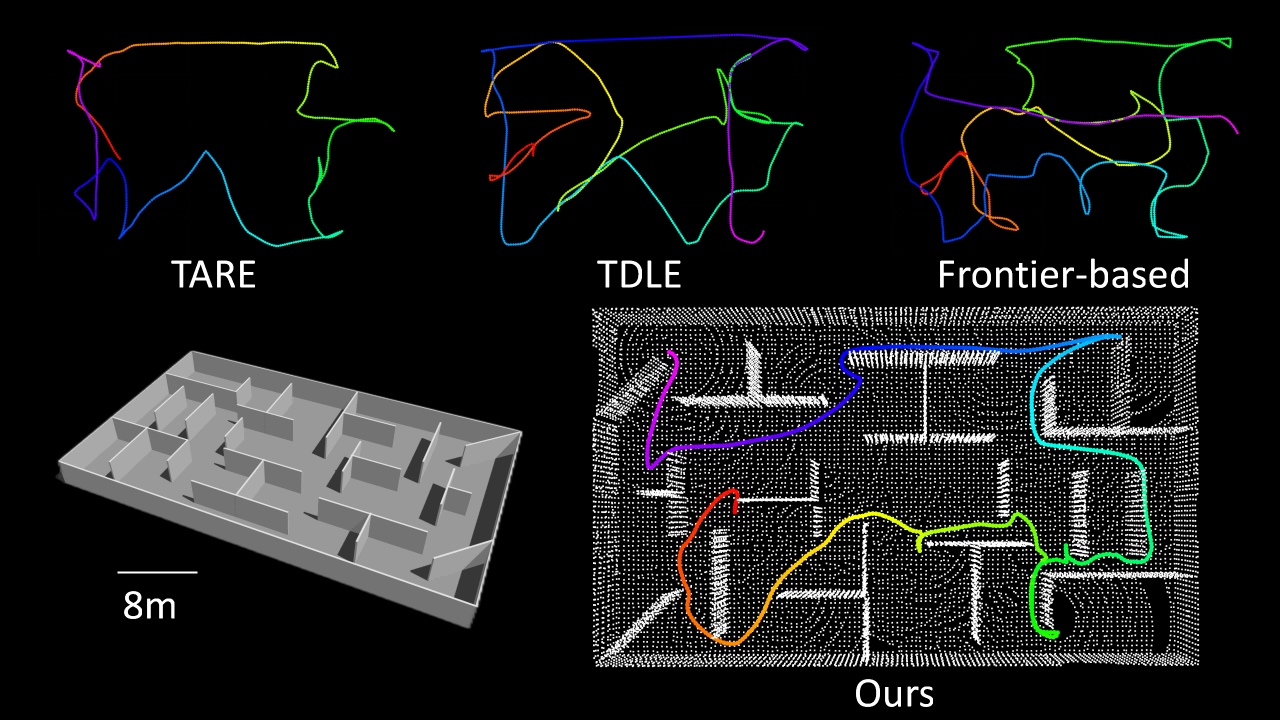}}
\hfill
\\ 
\subfigure[Office]{
    \includegraphics[width=6.5cm]{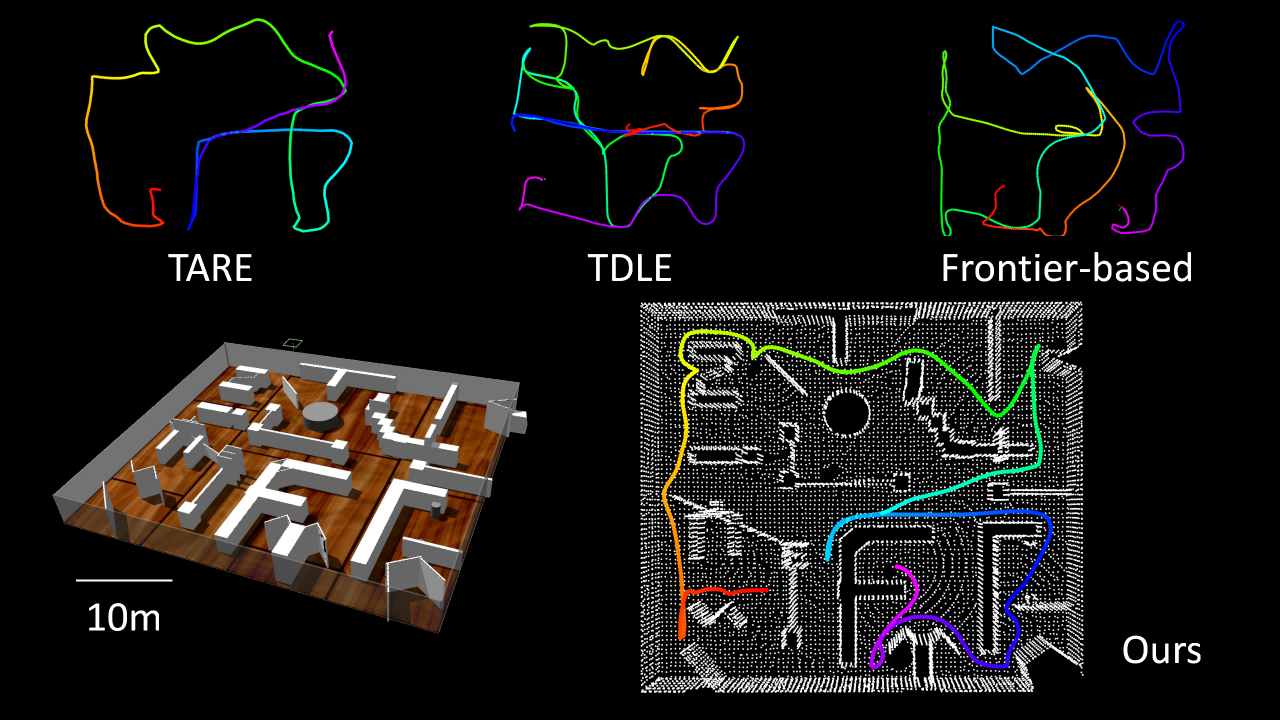}}
\hfill
\\
\subfigure[Indoor 1]{
    \includegraphics[width=6.5cm]{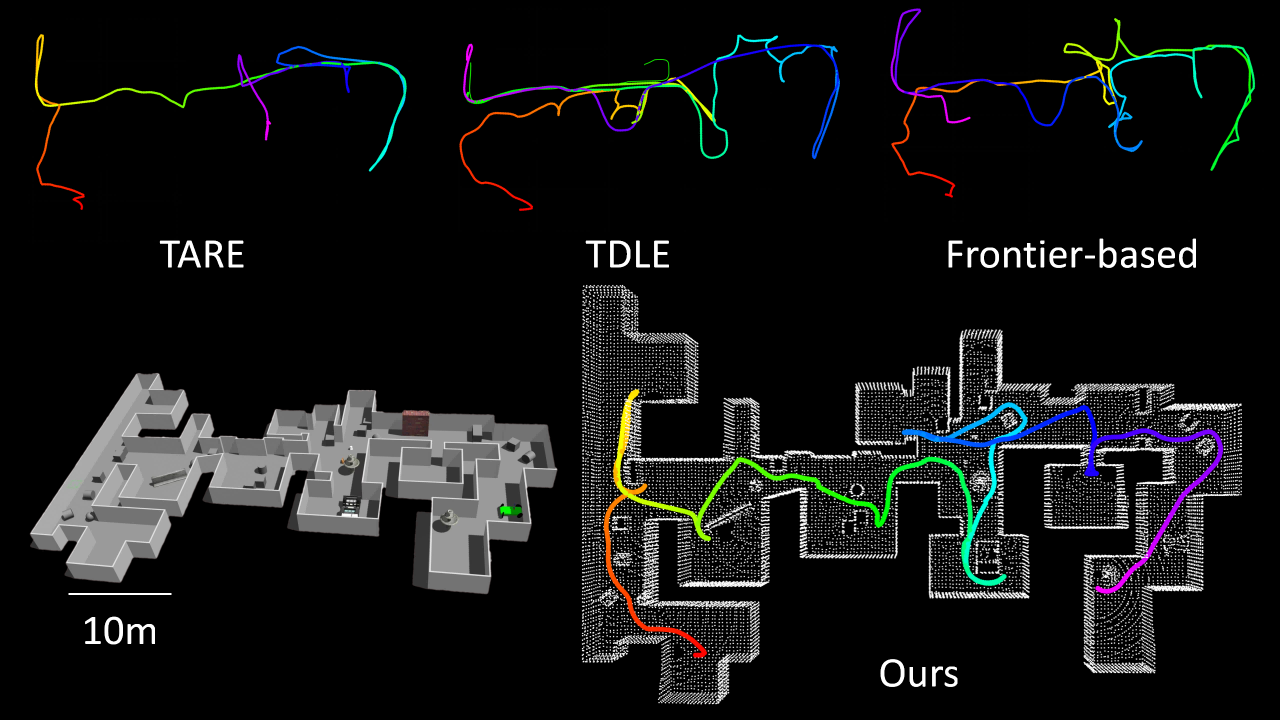}}
\\
\subfigure[Indoor 2]{
    \includegraphics[width=6.5cm]{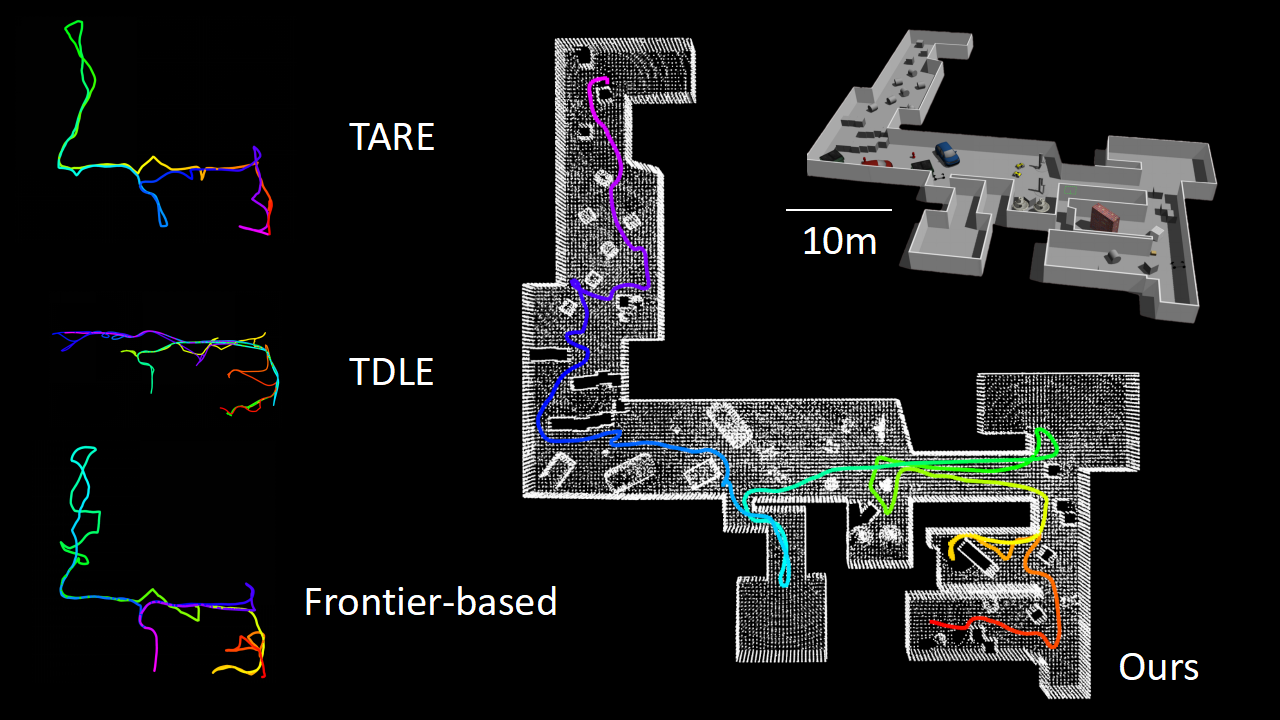}}
\caption{The result maps of our method and trajectories of all methods in four simulation scenes. The red end of the trajectory represents the start point, and the purple end of the trajectory represents the end point.}
\label{result map}
\end{figure}

The best trajectories for each method and their exploration results are depicted in Fig. \ref{result map}. From a purely path-based perspective, our approach exhibits minimal redundant paths and completes the entire exploration process seamlessly without any backtracking. Even in two indoor scenarios, our method sequentially passes through each room along the exploration path, ensuring comprehensive coverage with minimal details missed. On the other hand, the TARE exhibits several instances of backtracking as some rooms in the start and middle of the scene are left unexplored, necessitating a return to complete the exploration, and resulting in missed areas. Both the frontier-based method and the TDLE experience varying degrees of backtracking in all scenarios, leading to decreased exploration efficiency.

\begin{table}[h]
\caption{Simulation Results in Four Environments}
\begin{center}
\resizebox{\linewidth}{!}{
\begin{tabular}{c|c|c|c|c|c|c}
\hline
\hline
\multirow{2}{*}{\textbf{Scene}} & \multirow{2}{*}{\textbf{Method}} & \multicolumn{2}{c|}{\textbf{Distance}(\textit{m})} & \textbf{Exploration} & \multicolumn{2}{c}{\textbf{Time}(\textit{s})}\\
\cline{3-4}
\cline{6-7}
&  & \textit{Avg} & \textit{Std} &\textbf{ Rate }& \textit{Avg} & \textit{Std} \\
\hline
\multirow{4}{*}{\makecell[c]{Maze\\ ($700m^2$)}} & Frontier based & 242.2 & 42.5 & 2.89  & 345.2 & 57.7\\
& TDLE & 235.6 & 38.7 & 2.97  & 281.7 & 52.4\\
& TARE & 145.2 & 25.4 & 4.82  & 178.6 & \textbf{20.9} \\
& \textbf{Ours}  & \textbf{131.7} & \textbf{21.7} & \textbf{5.32}  & \textbf{171.3}& 28.3 \\
\hline

\multirow{4}{*}{\makecell[c]{Office\\ ($650m^2$)}} & Frontier based & 212.2 & 22.2 & 3.06 & 293.7 & 43.3\\
& TDLE & 294.5 & 33.8 & 2.21  & 375.6 & 46.7\\
& TARE & 191.6 & 29.8 & 3.35  & 230.4 & 35.4 \\
& \textbf{Ours}  & \textbf{135.9} & \textbf{22.5} & \textbf{4.78}  & \textbf{167.7} & \textbf{25.6} \\
\hline

\multirow{4}{*}{\makecell[c]{Indoor 1\\ ($850m^2$)}} & Frontier based & 270.6 & 42.1& 3.14  & 390.5 & 38.7\\
& TDLE & 290.5 & 35.6 & 2.93 & 509.8 & 33.8 \\
& TARE & 190.3 & 31.4 & 4.31  & 245.4 & \textbf{31.5} \\
& \textbf{Ours}  & \textbf{176.9} & \textbf{24.7} & \textbf{4.78}  & \textbf{223.9} & 32.7 \\
\hline
\multirow{4}{*}{\makecell[c]{Indoor 2\\ ($800m^2$)}} & Frontier based & 298.5 & 42.1& 2.68  & 405.5 & 42.5\\
& TDLE & 390.6 & 45.6 & 1.49 & 600 & 0$^{\mathrm{a}}$ \\
& TARE & 235.6 & \textbf{18.1} & 3.31  & 268.1 & \textbf{27.5} \\
& \textbf{Ours} &\textbf{197.5}  & 25.9 & \textbf{4.05}  & \textbf{208.2} & 29.8 \\
\hline
\hline
\multicolumn{7}{l}{$^{\mathrm{a}}$TDLE has not completed after 600s in the indoor 2 scene.}
\end{tabular}
}
\end{center}
\end{table}

Table 1 shows the specific experimental results and statistics. The results show that our method is able to explore more environmental information while consuming less traveling time and path length. Compared with the TARE, our method reduces the travel path by 7.0\% -29.1\% and the exploration time by 4.1\%-27.2\%, and improves the overall efficiency by 10.4\%-42.7\%. The hybrid frontier sampling method allows for the rapid extraction of potential frontiers from the environment while reducing the computation memory. Thus, the robot can immediately obtain effective frontier information when passing through unknown areas, ensuring timely exploration and planning. This is important because the small unexplored area has a huge impact on the whole exploration process, leading to a significant reduction in its later exploration efficiency.

\begin{figure}[!t]
\centering
\resizebox{\linewidth}{!}{
\subfigure[Maze]{
    \includegraphics[width=4cm]{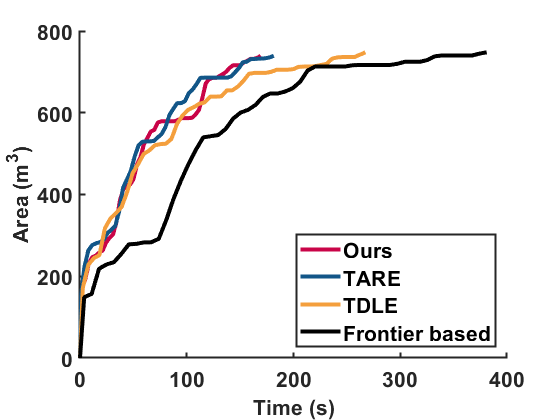}}
\subfigure[Office]{
    \includegraphics[width=4cm]{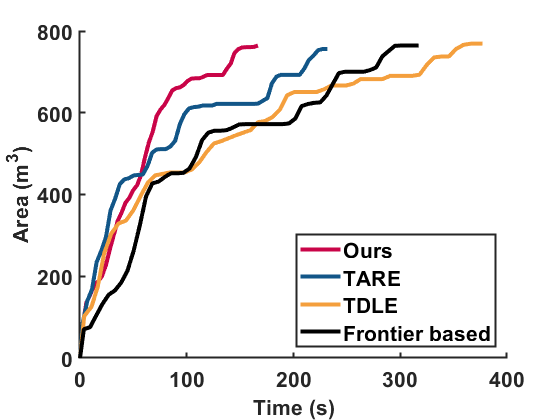}}
}
\\
\resizebox{\linewidth}{!}{
\subfigure[Indoor 1]{
    \includegraphics[width=4cm]{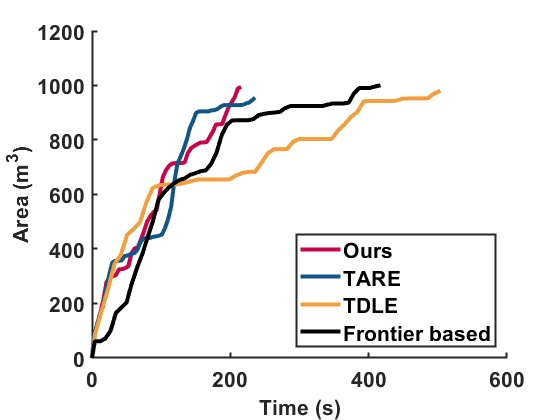}}
\subfigure[Indoor 2]{
    \includegraphics[width=4cm]{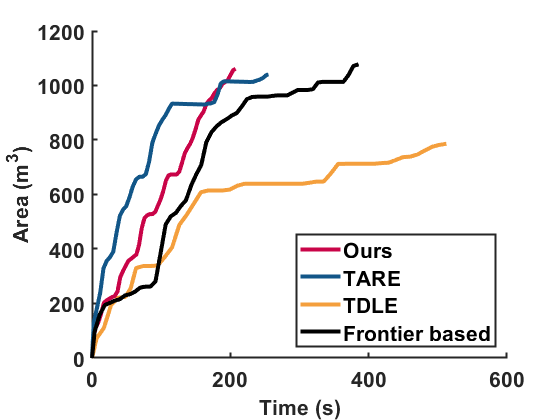}}
}
\caption{Comparison of the four methods in the exploration process.}
\label{area/time}
\end{figure}

Fig. \ref{area/time} visually shows the progress curves of each method during the exploration, and our method finishes the exploration first in all scenarios. Compared with the TDLE and frontier-based method, the exploration efficiency of our method is significantly higher. In two indoor scenes, the TARE makes fast exploration progress in the initial stage. After several experiments, the TARE exhibits a tendency to initially explore along the corridor, significantly accelerating progress in the earlier stages of the exploration. Then it proceeds to explore the smaller rooms, wherein the process of the accumulating exploration progress is relatively slower. In Fig. \ref{area/time} (c) and (d), the progress curves of the TARE initially grow rapidly, while the growth rate of curves slows down when returning to explore those rooms at the later stage. Our approach arranges the entry sequence of each subregion. The Eq. \ref{revenue}, encourages the robot to cover the whole exploration space with a short total path to avoid subsequent retrace, without the consideration of the information gain. As can be seen from the trajectories in Fig. \ref{result map} (c) and (d), our method can sequentially pass through each cell and region when exploring. Thus, our exploration progress initially grows at a relatively slow pace, but this approach helps us avoid the issue of unnecessary back-and-forth motion. Therefore, our method ends up traveling with a shorter path than TARE, and in the final stage of curves, it surpasses the TARE and completes the exploration earlier.

\begin{table}[h]
\caption{Exploration Completion for Each Method}
\begin{center}
\resizebox{8cm}{!}{
\begin{tabular}{c|c|c|c|c}
\hline
\hline
\multirow{2}{*}{\textbf{Method}} & \multicolumn{4}{c}{\textbf{Average exploration completion}}\\
\cline{2-5}
& \textbf{Maze} &\textbf{Office} &  \textbf{Indoor 1} &\textbf{Indoor 2} \\
\hline
Frontier based & 100\% & 100\% & 100\% & 100\% \\
TDLE & 100\% & 100\% & 98.4\% & 72.9\% \\
TARE & 98.4\% & 97.2\% & 96.5\% & 97.5\% \\
\textbf{Ours} & 100\% & 99.4\% & 99.5\% & 99.8\% \\
\hline
\hline
\end{tabular}
}
\end{center}
\end{table}

The average exploration completion (the ratio of the explored area to the total area) of each method is presented in Table 2. Our method achieves a higher degree of exploration completeness compared to the TARE, from 2.2\% to 3.0\%, while the TARE leaves some areas unexplored. The TDLE can detect the small unexplored corners, but it does not optimize the sampling method of the RRT tree, which leads to the subsequent turning back when some recent frontier points are not detected in time. Since our method also detects frontiers in the local map, it is able to quickly detect some small regions and corners that have not been explored and incorporate them into the planning process for the subsequent exploration.

\subsection{Real-World Experiment}
\begin{figure}[!t]
\centering
\subfigure[Maze 1]{
    \includegraphics[width=0.35\linewidth]{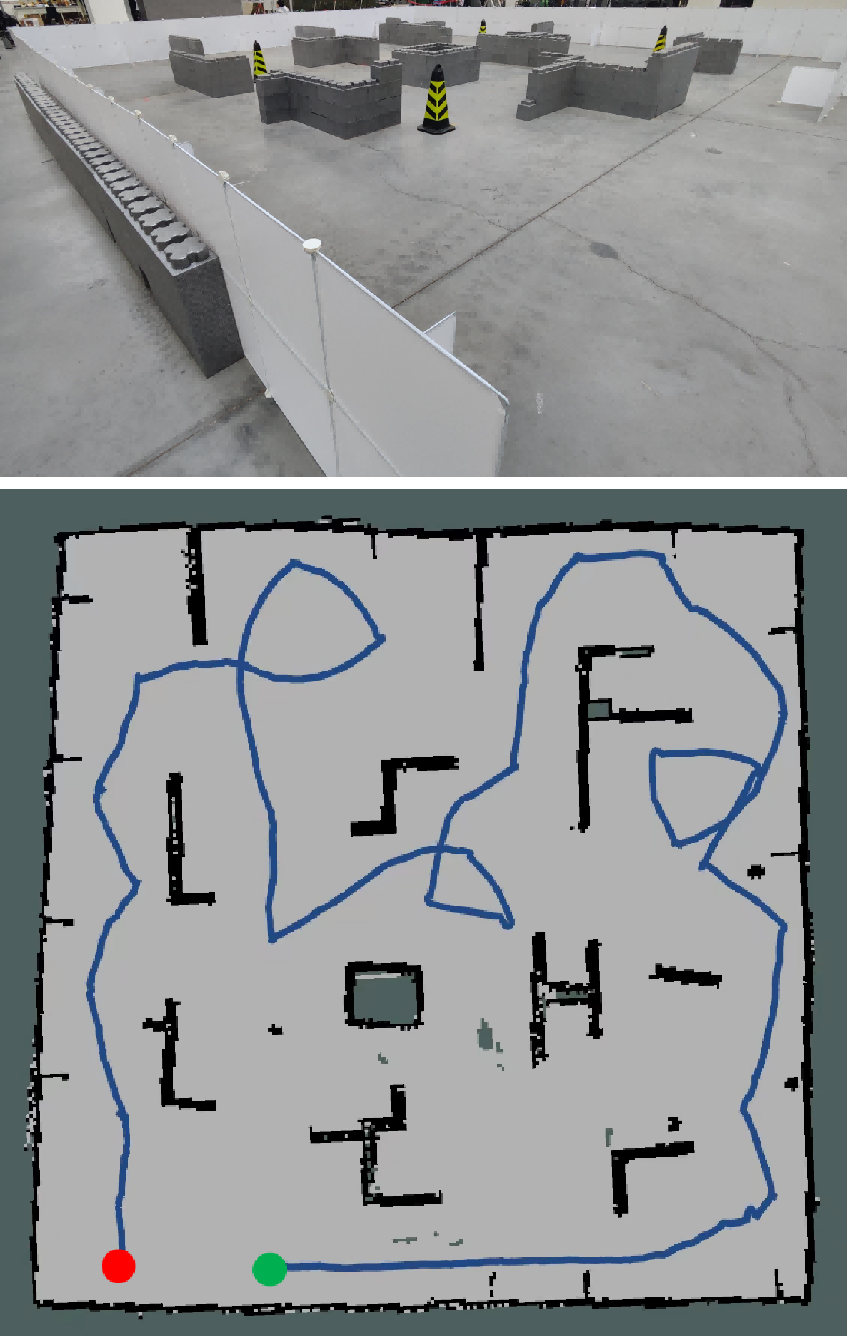}}
\subfigure[Maze 2]{
    \includegraphics[width=0.35\linewidth]{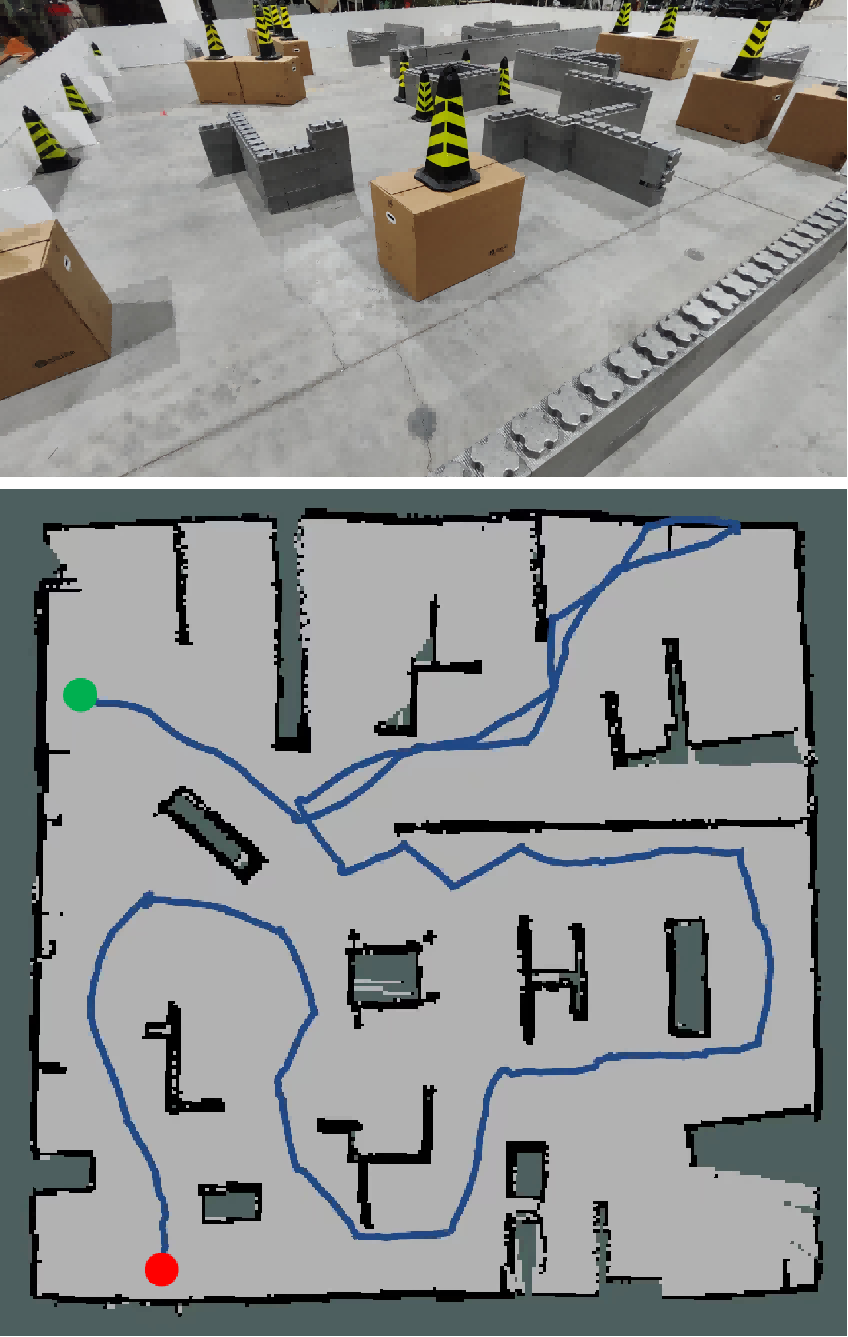}}
\\
\subfigure[Corridor]{
\includegraphics[width=0.6\linewidth]{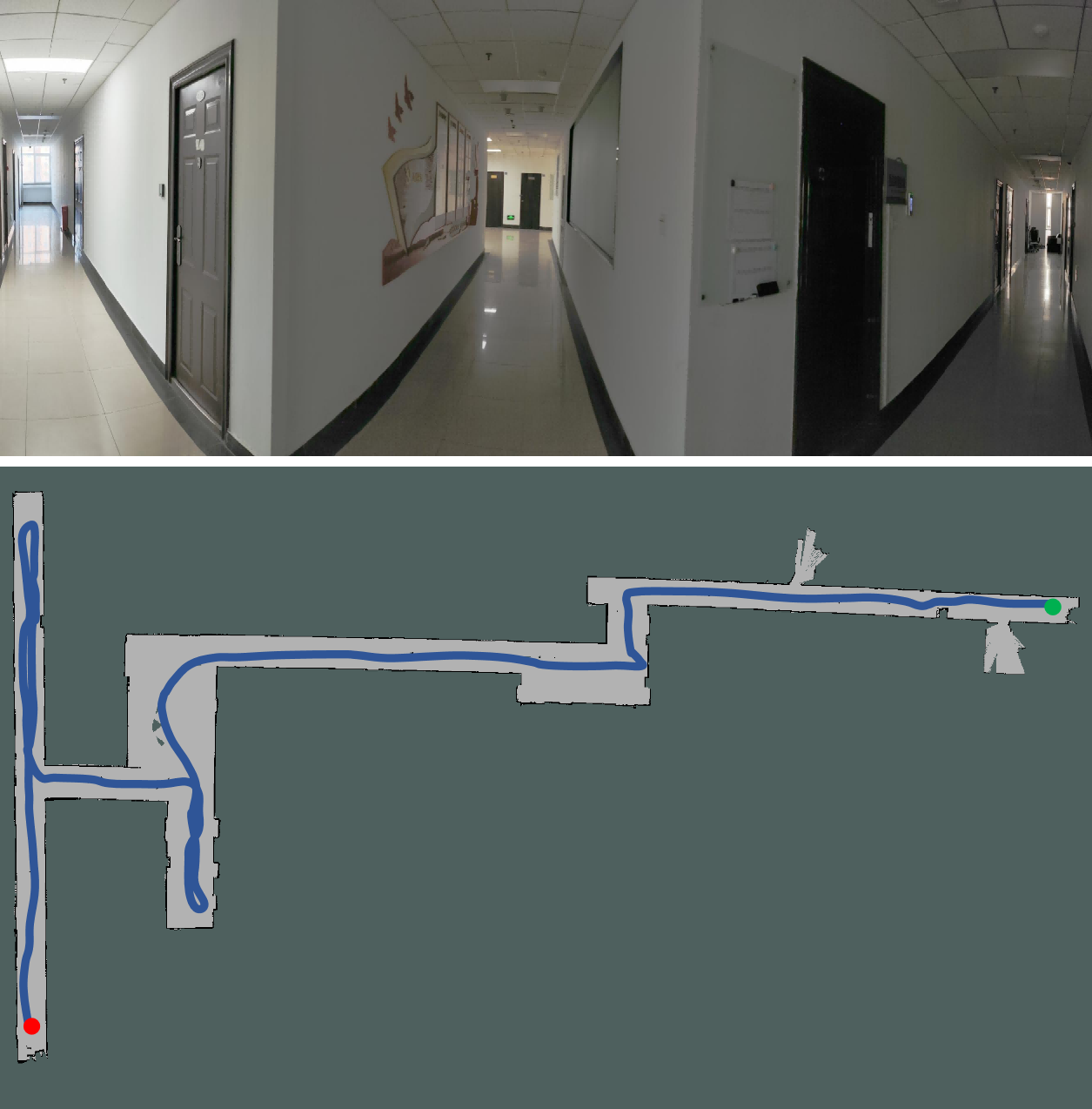}}
\caption{The real-world environments and exploration results. The red dot is the starting position and the green dot represents the end position.}
\label{real word}
\end{figure}

The real-world experiment is conducted in two human-made maze scenes and an indoor corridor scene. The maze scenes in Fig. \ref{real word} (a) and (b) are with the size of 10 m × 10 m, and different numbers of obstacles are set inside. Fig. \ref{real word} (c) is the corridor scene, with corridors criss-crossing each other. The Direct LiDAR Odometry \cite{dlo} is used as the SLAM module, and only relies on the LiDAR sensor for the localization. The maximum speed of the robot is set to 0.6 m/s.

Fig. \ref{real word} also shows the environments explored using the grid map created by Gmapping \cite{gmapping}, where the blue lines are the trajectories of the robot. The trajectories demonstrate the robot’s efficient exploration strategy without a redundant path. Even in the corridor with a more complex layout, the robot also visits each area in sequence, achieving an effective exploration in unknown environments.

\section{Conclusion}

We propose an efficient method to explore the unknown environment. Our method adopts the hybrid frontier sampling approach to rapidly extract frontier points by directly using the LiDAR data and the local map information. The hierarchical planning strategy is incorporated to drive the robot to explore the environment according to a path sequence determined by both global and local information. The simulation experiments demonstrate that the proposed method encompasses all regions while exhibiting significant advantages in terms of the traveling distance and the exploration time. The real-world experiments also further prove the effectiveness of our approach in realistic unknown environments.

\addtolength{\textheight}{-12cm}  

\bibliographystyle{IEEEtran}
\bibliography{paper}

\begin{thebibliography}{10}
\providecommand{\url}[1]{#1}
\csname url@samestyle\endcsname
\providecommand{\newblock}{\relax}
\providecommand{\bibinfo}[2]{#2}
\providecommand{\BIBentrySTDinterwordspacing}{\spaceskip=0pt\relax}
\providecommand{\BIBentryALTinterwordstretchfactor}{4}
\providecommand{\BIBentryALTinterwordspacing}{\spaceskip=\fontdimen2\font plus
\BIBentryALTinterwordstretchfactor\fontdimen3\font minus \fontdimen4\font\relax}
\providecommand{\BIBforeignlanguage}[2]{{%
\expandafter\ifx\csname l@#1\endcsname\relax
\typeout{** WARNING: IEEEtran.bst: No hyphenation pattern has been}%
\typeout{** loaded for the language `#1'. Using the pattern for}%
\typeout{** the default language instead.}%
\else
\language=\csname l@#1\endcsname
\fi
#2}}
\providecommand{\BIBdecl}{\relax}
\BIBdecl

\bibitem{yamauchi1997frontier}
B.~Yamauchi, ``A frontier-based approach for autonomous exploration,'' in \emph{Proceedings 1997 IEEE International Symposium on Computational Intelligence in Robotics and Automation CIRA'97.'Towards New Computational Principles for Robotics and Automation'}.\hskip 1em plus 0.5em minus 0.4em\relax IEEE, 1997, pp. 146--151.

\bibitem{perkasa2020improved}
D.~A. Perkasa and J.~Santoso, ``Improved frontier exploration strategy for active mapping with mobile robot,'' in \emph{2020 7th International Conference on Advance Informatics: Concepts, Theory and Applications (ICAICTA)}.\hskip 1em plus 0.5em minus 0.4em\relax IEEE, 2020, pp. 1--6.

\bibitem{image}
H.~Liang, Z.~Wang, and J.~Li, ``Robot autonomous frontier exploration strategy based on the idea of image processing.''

\bibitem{sensor_reading}
M.~Keidar and G.~A. Kaminka, ``Efficient frontier detection for robot exploration,'' \emph{The International Journal of Robotics Research}, vol.~33, no.~2, pp. 215--236, 2014.

\bibitem{huang2023fael}
J.~Huang, B.~Zhou, Z.~Fan, Y.~Zhu, Y.~Jie, L.~Li, and H.~Cheng, ``Fael: Fast autonomous exploration for large-scale environments with a mobile robot,'' \emph{IEEE Robotics and Automation Letters}, vol.~8, no.~3, pp. 1667--1674, 2023.

\bibitem{gao2018improved}
W.~Gao, M.~Booker, A.~Adiwahono, M.~Yuan, J.~Wang, and Y.~W. Yun, ``An improved frontier-based approach for autonomous exploration,'' in \emph{2018 15th International Conference on Control, Automation, Robotics and Vision (ICARCV)}.\hskip 1em plus 0.5em minus 0.4em\relax IEEE, 2018, pp. 292--297.

\bibitem{deep1}
F.~Niroui, K.~Zhang, Z.~Kashino, and G.~Nejat, ``Deep reinforcement learning robot for search and rescue applications: Exploration in unknown cluttered environments,'' \emph{IEEE Robotics and Automation Letters}, vol.~4, no.~2, pp. 610--617, 2019.

\bibitem{deep2}
A.~Feng, Y.~Xie, Y.~Sun, X.~Wang, B.~Jiang, and J.~Xiao, ``Efficient autonomous exploration and mapping in unknown environments,'' \emph{Sensors}, vol.~23, no.~10, p. 4766, 2023.

\bibitem{deep3}
W.-C. Lee, M.~C. Lim, and H.-L. Choi, ``Extendable navigation network based reinforcement learning for indoor robot exploration,'' in \emph{2021 IEEE International Conference on Robotics and Automation (ICRA)}.\hskip 1em plus 0.5em minus 0.4em\relax IEEE, 2021, pp. 11\,508--11\,514.

\bibitem{deep4}
L.~Schmid, M.~N. Cheema, V.~Reijgwart, R.~Siegwart, F.~Tombari, and C.~Cadena, ``Sc-explorer: Incremental 3d scene completion for safe and efficient exploration mapping and planning,'' \emph{arXiv preprint arXiv:2208.08307}, 2022.

\bibitem{deep5}
K.~Leong, ``Reinforcement learning with frontier-based exploration via autonomous environment,'' \emph{arXiv preprint arXiv:2307.07296}, 2023.

\bibitem{holz2010evaluating}
D.~Holz, N.~Basilico, F.~Amigoni, and S.~Behnke, ``Evaluating the efficiency of frontier-based exploration strategies,'' in \emph{ISR 2010 (41st International Symposium on Robotics) and ROBOTIK 2010 (6th German Conference on Robotics)}.\hskip 1em plus 0.5em minus 0.4em\relax VDE, 2010, pp. 1--8.

\bibitem{nbvp}
A.~Bircher, M.~Kamel, K.~Alexis, H.~Oleynikova, and R.~Siegwart, ``Receding horizon" next-best-view" planner for 3d exploration,'' in \emph{2016 IEEE international conference on robotics and automation (ICRA)}.\hskip 1em plus 0.5em minus 0.4em\relax IEEE, 2016, pp. 1462--1468.

\bibitem{tdle}
X.~Zhao, C.~Yu, E.~Xu, and Y.~Liu, ``Tdle: 2-d lidar exploration with hierarchical planning using regional division,'' in \emph{2023 IEEE 19th International Conference on Automation Science and Engineering (CASE)}, 2023, pp. 1--6.

\bibitem{tare}
C.~Cao, H.~Zhu, H.~Choset, and J.~Zhang, ``Tare: A hierarchical framework for efficiently exploring complex 3d environments.'' in \emph{Robotics: Science and Systems}, vol.~5, 2021.

\bibitem{dsvp}
H.~Zhu, C.~Cao, Y.~Xia, S.~Scherer, J.~Zhang, and W.~Wang, ``Dsvp: Dual-stage viewpoint planner for rapid exploration by dynamic expansion,'' in \emph{2021 IEEE/RSJ International Conference on Intelligent Robots and Systems (IROS)}.\hskip 1em plus 0.5em minus 0.4em\relax IEEE, 2021, pp. 7623--7630.

\bibitem{fuel}
B.~Zhou, Y.~Zhang, X.~Chen, and S.~Shen, ``Fuel: Fast uav exploration using incremental frontier structure and hierarchical planning,'' \emph{IEEE Robotics and Automation Letters}, vol.~6, no.~2, pp. 779--786, 2021.

\bibitem{gdae}
R.~Cimurs, I.~H. Suh, and J.~H. Lee, ``Goal-driven autonomous exploration through deep reinforcement learning,'' \emph{IEEE Robotics and Automation Letters}, vol.~7, no.~2, pp. 730--737, 2021.

\bibitem{dtw}
E.~Keogh and C.~A. Ratanamahatana, ``Exact indexing of dynamic time warping,'' \emph{Knowledge and information systems}, vol.~7, pp. 358--386, 2005.

\bibitem{informed}
R.~Cimurs, I.~H. Suh, and J.~H. Lee, ``Information-based heuristics for learned goal-driven exploration and mapping,'' in \emph{2021 18th International Conference on Ubiquitous Robots (UR)}.\hskip 1em plus 0.5em minus 0.4em\relax IEEE, 2021, pp. 571--578.

\bibitem{frontier}
J.~Or{\v{s}}uli{\'c}, D.~Mikli{\'c}, and Z.~Kova{\v{c}}i{\'c}, ``Efficient dense frontier detection for 2-d graph slam based on occupancy grid submaps,'' \emph{IEEE Robotics and Automation Letters}, vol.~4, no.~4, pp. 3569--3576, 2019.

\bibitem{rrt_exploration}
H.~Umari and S.~Mukhopadhyay, ``Autonomous robotic exploration based on multiple rapidly-exploring randomized trees,'' in \emph{2017 IEEE/RSJ International Conference on Intelligent Robots and Systems (IROS)}, 2017, pp. 1396--1402.

\bibitem{cao2022autonomous}
C.~Cao, H.~Zhu, F.~Yang, Y.~Xia, H.~Choset, J.~Oh, and J.~Zhang, ``Autonomous exploration development environment and the planning algorithms,'' in \emph{2022 International Conference on Robotics and Automation (ICRA)}.\hskip 1em plus 0.5em minus 0.4em\relax IEEE, 2022, pp. 8921--8928.

\bibitem{dlo}
K.~Chen, B.~T. Lopez, A.-a. Agha-mohammadi, and A.~Mehta, ``Direct lidar odometry: Fast localization with dense point clouds,'' \emph{IEEE Robotics and Automation Letters}, vol.~7, no.~2, pp. 2000--2007, 2022.

\bibitem{gmapping}
G.~Grisetti, C.~Stachniss, and W.~Burgard, ``Improved techniques for grid mapping with rao-blackwellized particle filters,'' \emph{IEEE transactions on Robotics}, vol.~23, no.~1, pp. 34--46, 2007.

\end{thebibliography}
\end{document}